\pdfoutput=1
\documentclass[11pt]{article}
\usepackage[table,usenames,dvipsnames]{xcolor}

\usepackage[final]{acl}
\usepackage{tabularx}
\usepackage{comment}
\usepackage{placeins}
\usepackage{stfloats} 
\usepackage{placeins}
\usepackage{float} % add this line
\usepackage{tabu} 
\usepackage{multirow}
\usepackage{booktabs}
\usepackage{hyperref}

\usepackage{etoolbox} % Required for modifying environments

\renewenvironment{quote}
  {\list{}{\rightmargin=0.2cm \leftmargin=0.2cm}%
   \item\relax}
  {\endlist}
\usepackage{caption}
\usepackage{times}
\usepackage{latexsym}
\usepackage{graphicx}
\usepackage{amsmath}
\usepackage{threeparttable}
\usepackage{paralist}
\usepackage{amssymb}
\usepackage{subcaption}
\usepackage{float}

\usepackage{tikz}
\usepackage{array}
\usepackage{tabularx}
\usepackage{cellspace}
\usetikzlibrary{shapes,arrows,positioning}
\definecolor{ao(english)}{rgb}{0.0, 0.5, 0.0}
\definecolor{brass}{rgb}{0.6, 0.8, 0.2}
\definecolor{chromeyellow}{rgb}{1.0, 0.65, 0.0}
\definecolor{crimson}{rgb}{0.86, 0.08, 0.26}

\usepackage{titlesec}

\titlespacing*{\paragraph}{%
  0pt}{%              left margin
  0.3\baselineskip}{% space before (vertical)
  1em}%               space after (horizontal)

\usepackage{times}
\usepackage{latexsym}
\usepackage{float} % add this line
\usepackage[T1]{fontenc}
\usepackage[utf8]{inputenc}

\usepackage{microtype}

\usepackage{inconsolata}
\usepackage{paralist}
\usepackage{pgfplots}
\pgfplotsset{width=10cm,compat=1.17}
\usepackage{pgf-pie}
\usepackage{footnote}
\makesavenoteenv{tabular}
\makesavenoteenv{table}
\usepackage[most]{tcolorbox}
\usepackage{enumitem}
\usepackage[absolute,showboxes]{textpos} % Add `showboxes` for debugging
\textblockorigin{0pt}{0pt} % Set origin to top left corner
\TPshowboxestrue % Show text block boxes, useful for debugging
\TPshowboxesfalse % Turn off show boxes for final output
\usetikzlibrary{pgfplots.polar, positioning}

\newtcolorbox{mybox}[2][]{enhanced, 
    colback=blue!2!white, colframe=blue!75!black, 
    fonttitle=\bfseries, coltitle=white, title=#2,
    sharp corners=south, rounded corners=north,
    boxrule=0.4mm, boxsep=5pt,
    borderline={0.3mm}{0.3mm}{gray!50!black},
    drop shadow={white, opacity=0.5, xshift=2.5mm, yshift=-2.5mm},
    attach boxed title to top left={yshift=-3mm, xshift=5mm},
    boxed title style={colback=blue!75!black, sharp corners},
    #1}

\definecolor{pastelgreen}{rgb}{0.75, 1.0, 0.85}
\definecolor{pastelblue}{rgb}{0.85, 0.85, 1.0}
\definecolor{pastelpink}{rgb}{1.0, 0.85, 0.9}
\definecolor{pastelyellow}{rgb}{1.0, 1.0, 0.8}
\definecolor{pastelpurple}{HTML}{D1CCEA} % Adjusted in HTML notation for finer control
\definecolor{pastelorange}{rgb}{1.0, 0.9, 0.75}
\definecolor{pastelred}{rgb}{1.0, 0.8, 0.8}
\definecolor{pastelgray}{rgb}{0.9, 0.9, 0.9}
\definecolor{pastelcyan}{rgb}{0.8, 0.95, 1.0}
\definecolor{pastelbrown}{rgb}{0.95, 0.85, 0.75}
\definecolor{pastelpurplebox}{rgb}{0.75, 0.58, 0.89}

\newcolumntype{L}[1]{>{\raggedright\let\newline\\\arraybackslash\hspace{0pt}}m{#1}}
\newcolumntype{C}[1]{>{\centering\let\newline\\\arraybackslash\hspace{0pt}}m{#1}}
\newcolumntype{R}[1]{>{\raggedleft\let\newline\\\arraybackslash\hspace{0pt}}m{#1}}

\newcommand{\corpusname}[0]{\textsc{MirrorStories}}

\title{%
    \begin{textblock*}{1cm}[0,0](2.7cm,2.1cm)  % {block width} (coords) 
    \includegraphics[height=1.5cm]{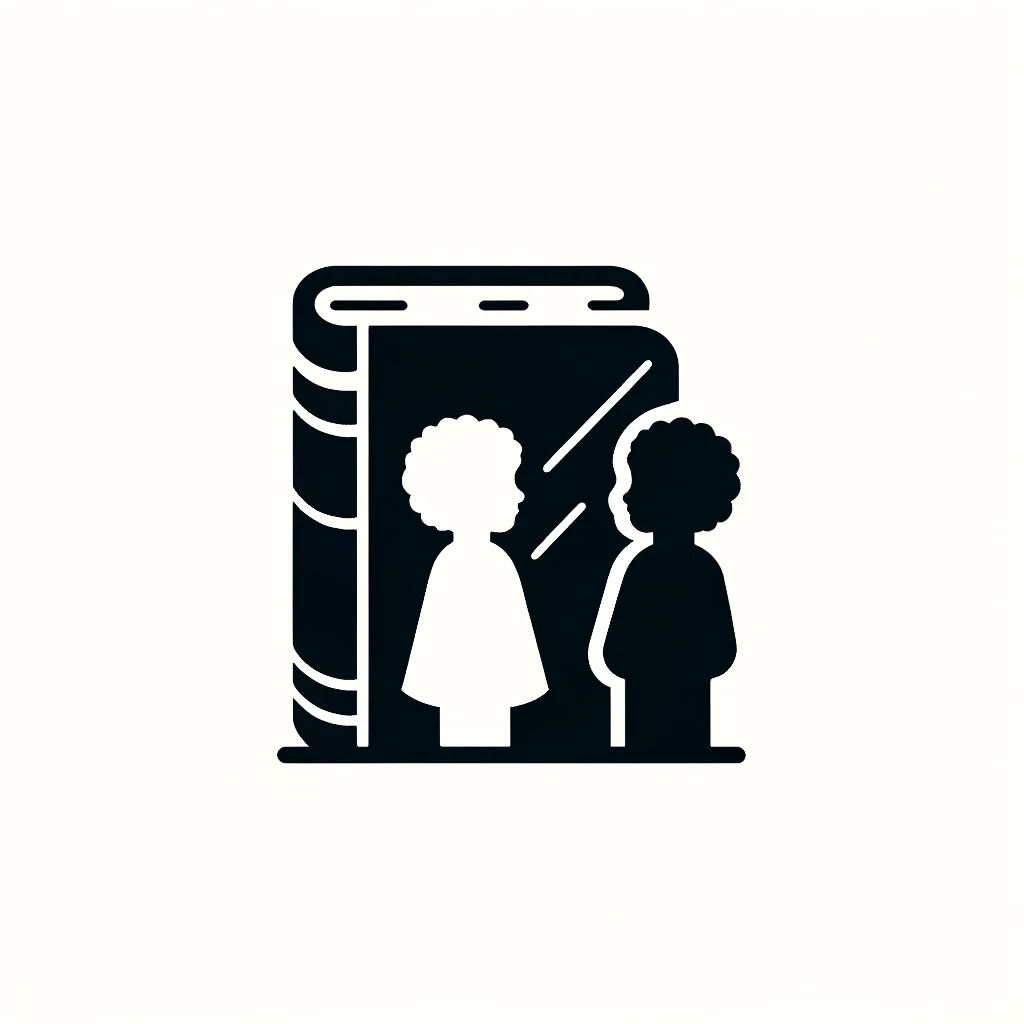}
    \end{textblock*}
    \corpusname{:} Reflecting Diversity through Personalized \\Narrative Generation with Large Language Models
}

\author{%
  Sarfaroz Yunusov\textnormal{,} 
  Hamza Sidat\textnormal{,}  \textnormal{and}  
  Ali Emami\\
  Brock University, St. Catharines, Canada \\
\texttt{\{zw22fi, hs18so, aemami\}@brocku.ca} \\
}

\begin{document}

\maketitle
\begin{abstract}

This study explores the effectiveness of Large Language Models (LLMs) in creating personalized ``mirror stories'' that reflect and resonate with individual readers' identities, addressing the significant lack of diversity in literature. We present \corpusname{}, a corpus of 1,500 personalized short stories generated by integrating elements such as name, gender, age, ethnicity, reader interest, and story moral. We demonstrate that LLMs can effectively incorporate diverse identity elements into narratives, with human evaluators identifying personalized elements in the stories with high accuracy. Through a comprehensive evaluation involving 26 diverse human judges, we compare the effectiveness of \corpusname{} against generic narratives. We find that personalized LLM-generated stories not only outscore generic human-written and LLM-generated ones across all metrics of engagement (with average ratings of 4.22 versus 3.37 on a 5-point scale), but also achieve higher textual diversity while preserving the intended moral. We also provide analyses that include bias assessments and a study on the potential for integrating images into personalized stories.\footnote{Interactive web application and corpus are publicly available at \href{https://www.mirrorstories.me}{mirrorstories.me}.}

\end{abstract}

\section{Introduction}

\begin{quote}
\textit{``There is no greater agony than bearing an untold story inside you.''} — Maya Angelou
\end{quote}

% 1) We need to include extended motivation and concrete applications of personalized book. I thought it might be a nice place to have them here. 

\textit{Mirror books} are stories that reflect the reader's identity, culture, and experiences, serving to engage, validate, and empower individuals \citep{bishop1990mirrors}. Such books are crucial in educational settings, fostering a sense of belonging and self-understanding through diverse narratives \citep{fleming2016more}, while also improving engagement and comprehension \citep{inbook, https://doi.org/10.1002/trtr.2139}. Beyond education, personalized narratives have shown potential in various fields, including health communication and marketing, where they enhance patient understanding and adherence, and strengthen emotional connections between brands and consumers \citep{202402.1709, article2}.

Despite the profound need for these personalized narratives, there is a noticeable underrepresentation of non-white minority groups in literature \citep{ccbc2021} relative to their population size, detailed in Appendix Figure \ref{fig:diversity}. The gap in cultural representation highlights the need for more inclusive narratives that reflect diverse reader identities, enhance empathy, and promote cultural awareness \citep{hoytt2022impact}. Diversity in literature can lead to improved innovation and a broader consideration of ideas, ultimately enriching the reading experience for all \citep{article_phil}.

%Studies also reveal that when children do not see themselves represented in the books they read, it can significantly impact their self-esteem and sense of self-worth, especially for children from underrepresented groups \citep{henderson2020take}. 

\begin{figure*}[ht]
    \centering
    \includegraphics[width=0.9\textwidth]{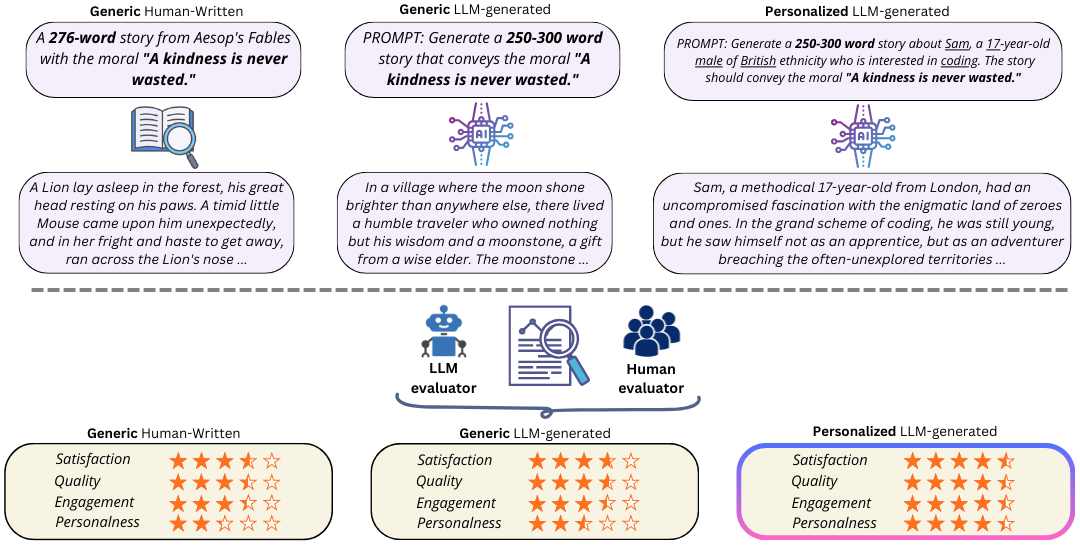}
    \caption{Generation and evaluation process for human-written generic, LLM-generated generic, and LLM-generated personalized narratives}
    \label{fig:cover}
    \vspace{-2mm}
\end{figure*}

Advancements in natural language processing, particularly through the development of LLMs like GPT-4, PaLM, and LLaMA have introduced the potential to address these gaps on a large scale \citep{openai2023gpt4,chowdhery2022palm,touvron2023llama}. LLMs excel in generating human-like text and adapting content to various contextual needs \citep{brown2020language, zhao2023survey}.

Recent studies have investigated LLMs' capabilities in expressing personality within generated content \citep{li2024tailoring, jiang2024personallm} and developing methods to induce and edit personality expressions in LLM outputs \citep{jiang2023evaluating, li2024tailoring, mao2024editing}. New benchmarks have also been released to assess personality traits in LLM outputs \citep{jiang2023evaluating, wang2024incharacter}. However, there remains a gap in research concerning whether LLMs can generate content that incorporates identity traits and faithfully mirrors the diverse identities of a global readership.

%Many startups, such as Riiid\footnote{https://riiid.com/} and Sana Labs\footnote{https://sanalabs.com/}, have also emerged to explore further the aspects of personalization in LLMs, leveraging these capabilities to provide users with personalized educational content and AI tutors on their rapidly growing platforms. 

Our study addresses this gap by exploring the potential of LLMs to create mirror stories—narratives that genuinely reflect and resonate with the identities of individual readers. We present a framework that evaluates the effectiveness of LLM-generated mirror stories in comparison to traditional narratives, assessing their impact on engagement, satisfaction, and the perception of personal relevance (see Figure~\ref{fig:cover}). Our contributions are three-fold:
\begin{compactenum}
\item We release \corpusname{}, a corpus of 1,500 personalized short stories generated by integrating elements such as name, gender, age, ethnicity, reader interest, and moral of the story. We demonstrate that LLMs can effectively incorporate identity elements into narratives, with human evaluators identifying them in the stories with high accuracy.
\item Through a comprehensive evaluation involving 26 diverse human judges, personalized LLM-generated stories consistently outperform both generic human-written and LLM-generated stories across all engagement metrics, with a significantly higher average rating.
\item We present analyses that assess text diversity, coherence, and moral comprehension across each story type, and examine biases exhibited by LLMs when evaluating personalized narratives. We also explore the potential of integrating images and incorporate \corpusname{} into an interactive \textbf{\href{www.mirrorstories.me}{web application}} where users can browse and generate stories.
\end{compactenum}

\section{\corpusname{}}

\subsection{Overview}

\corpusname{} is a corpus designed to evaluate the ability of LLMs to generate both generic and personalized short narratives based on predefined morals and identity elements. Each dataset instance consists of a moral (e.g., ``Kindness is never wasted'') guiding the narrative's tone and a set of identity elements (name, age, gender, ethnicity, and personal interest) to personalize the story. Specifically, the dataset includes a human-written and an LLM-generated generic story, both of which do not incorporate specific identity elements, and an LLM-generated personalized story that includes these elements to enhance relevance and engagement. Appendix \ref{sec:subsectionA4} provides a detailed example of the dataset structure.

% Each dataset instance consists of a moral (e.g., "Kindness is never wasted") guiding the narrative's tone, a set of identity traits (name, age, gender, ethnicity, and personal interest) to personalize the story, a human-written generic story and an LLM-generated generic story, both adhering to the same moral but without incorporating specific identity traits, and an LLM-generated personalized story integrating both the moral and the identity elements to enhance relevance and engagement.

\iffalse \paragraph{Example Instance}
Consider the moral ``Kindness is never wasted.'' For a generic story, the narrative might revolve around an abstract act of kindness in a village. For the personalized version, the story could focus on Priya, a 17-year-old from London with an interest in coding, where the narrative integrates her background and interests, showing how his coding skills helped someone in need, weaving the moral into a personalized context.
\fi

\begin{figure*}[t]
    \centering
    \includegraphics[width=\textwidth]
    {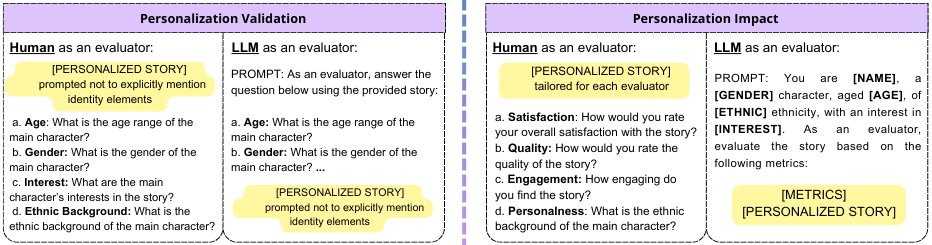}
    
    \caption{Illustration demonstrating the personalization validation and impact processes}
    \label{fig:personalization_test_story}
    \vspace{-1mm}
\end{figure*}

\subsection{Dataset Collection}

\paragraph{Human-written Stories \& Morals}
\corpusname{} incorporates human-written stories derived from Aesop's fables \cite{wier1890aesop}\footnote{Scraped from \url{https://read.gov/aesop/001.html}}, a well-known collection famous for its clear narrative structure and explicit moral conclusions. The morals serve as guides for generating both generic and personalized stories. The complete list of morals is provided in Appendix \ref{sec:subsectionA4} Table~\ref{tab:dataset_values}.
\paragraph{Identities}
Identity traits such as name, age, gender, ethnic background, and interests are included to personalize the narratives. We drew from 123 unique ethnic backgrounds, 124 diverse interests, and 28 distinct morals. The complete set of identities is provided in Appendix \ref{sec:subsectionA4} Table~\ref{tab:dataset_values}.
\paragraph{Generic \& Personalized LLM-Generated Stories}
Generic stories are generated focusing solely on the moral, while personalized stories additionally integrate the specified identity elements. For the specific prompts, refer to Figure \ref{fig:cover}.  GPT-4 (ver. 0613), Claude-3 Sonnet\footnote{\url{https://www.anthropic.com/claude}}, and Gemini 1.5 Flash \cite{reid2024gemini} were used, each responsible for generating one-third of the narratives. 

\corpusname{} comprises 1,500 narratives with an almost even split between male and female characters. The dataset spans a broad age range from 10 to 60 years. Detailed illustrations of the distributions are in Appendix \ref{sec:appendix_A1} Figures \ref{fig:gender_mirror} and \ref{fig:age_mirror}.

\section{Experiments}

We conducted two experiments to assess the effectiveness of personalization in LLM-generated stories: \textit{Personalization Validation}, which validates the integration of identity elements within the narratives, and  \textit{Personalization Impact}, which assesses the impact of these narratives on user engagement, comprehension, satisfaction, and personalness.%Figure~\ref{fig:personalization_test_story} provides an overview of the two experiments.

\paragraph{Prompts}
In both experiments, personalized prompts incorporating identity elements were used to generate personalized stories. For Personalization Validation, these elements were specifically asked not to be stated explicitly, to test their seamless integration into the narrative. In the Personalization Impact experiment, personal elements were aligned with those of 26 human evaluators, ensuring that each story was tailored to evaluators. Figure~\ref{fig:cover} and~\ref{fig:personalization_test_story} provide detailed structures of the prompts for both generation and evaluation.

% For both experiments, personalized prompts that include identity elements such as name, age, gender, ethnic background, and interests were used. These traits were carefully matched with those of 26 human annotators to ensure that each story was personalized to mirror the specific reader evaluating it. In the Personalization Validation, evaluators determined how well these personalized elements were integrated into each story. Detailed structures of these prompts are shown in Figure~\ref{fig:personalization_test_story}.

\paragraph{Human Evaluation}
In both experiments, the same 26 human evaluators—all students from diverse disciplines—were tasked with evaluating the narratives (for demographic details, see Appendix Figure \ref{fig:annotators_demongraphics}). For the Personalization Validation, they answered a structured questionnaire for a random sample of 30 stories to identify the personalized elements. In the Personalization Impact test, each evaluator reviewed a human-written, generic LLM-generated, and personalized LLM-generated story, with the personalized LLM-generated story specifically tailored to reflect their personal identity. They provided feedback on all three story types, rating them on satisfaction, quality, engagement, and personalness. The detailed questionnaire is provided in Appendix \ref{appendix_A2}.

% In the Personalization Impact test, each evaluator assessed human-written, generic LLM-generated and personalized LLM-generated story, which was specifically crafted to reflect their personal identity, providing feedback on how well the stories reflected their own experiences and traits. The questionnaire is provided in Appendix \ref{appendix_A2}.

\paragraph{Models}
   GPT-4 (ver. 0613, temperature 0.4) was used as an evaluator in both experiments. Initially, it assessed the integration of personalized elements. Later, it was used to evaluate the stories for satisfaction, quality, engagement, and personalness, with a sample of the evaluation process and prompts provided in Appendix Figure \ref{GPT4_Evaluator}. GPT-4 was chosen for its increasing adoption as an evaluator across domains \citep{Gilardi_2023, tarkka-etal-2024-automated, 10400468}, with potential advantages such as scalability, cost-efficiency, and consistency. %We aim to explore whether GPT-4 can accurately simulate human judgment in creative tasks, particularly in evaluating stories. Should a strong correlation be found, it would be further used to assess a larger volume of stories, enabling deeper insights that go beyond limitations of 26 human evaluators. 

\section{Results}

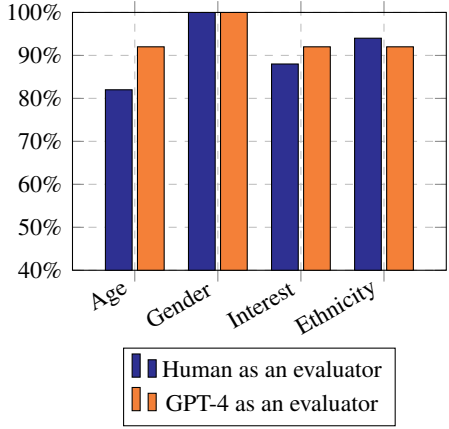
\begin{figure}[t]
\centering
\begin{tikzpicture}
\begin{axis}[
    % title=\textbf{\small Accuracy},
    ylabel style={font=\fontsize{8pt}{10pt}\selectfont},
    xlabel style={font=\fontsize{8pt}{10pt}\selectfont},
    ybar,
    enlarge x limits=0.25,
    xtick=data,
    xticklabels={Age, Gender, Interest, Ethnicity},
    x tick label style={rotate=30,anchor=east},
    xticklabel style={font=\footnotesize}, 
    yticklabel style={font=\footnotesize}, 
    legend style={at={(0.5,-0.30)},anchor=north,font=\small}, 
    grid=major, 
    grid style={dashed},
    ymin=40, ymax=100, 
    width=6.5cm,
    height=5cm,
    ytick={40,50,60,70,80,90,100},
    yticklabels={40\%,50\%,60\%,70\%,80\%,90\%,100\%}
]

\addplot+[ybar, fill=Blue, draw=black] coordinates {
    (1,82) (2,100) (3,88) (4,94)
};
\addlegendentry{Human as an evaluator}

\addplot+[ybar, fill=Orange, draw=black] coordinates {
    (1,92) (2,100) (3,92) (4,92)
};
\addlegendentry{GPT-4 as an evaluator}

\end{axis}
\end{tikzpicture}
\\
\caption{Accuracy of human and LLM evaluators in identifying identity elements in the story}
\label{fig:personalization_test_story_results}
\end{figure}

\paragraph{Are \corpusname{} personalized?}
The effectiveness of personalization in \corpusname{} is evident from the high accuracy rates in identifying identity elements by both human and LLM evaluators. As shown in Figure~\ref{fig:personalization_test_story_results}, human evaluators were particularly adept at identifying gender and ethnicity with accuracies at 100\% and 94\%, respectively. Similarly, GPT-4 showed robust performance, matching or exceeding human accuracy in all categories, which confirms the high level of personalization achieved in the narratives.

Personalized LLM-generated stories also effectively incorporate both the provided moral and the reader's interests, with a stronger emphasis on the moral. To demonstrate this, we used BERTopic \cite{grootendorst2022bertopic} for topic modeling to identify the top five terms for each story. We then calculated cosine similarity using Word2Vec embeddings\footnote{\textit{word2vec-google-news-300}} \citep{NIPS2013_9aa42b31} between these top terms and the provided interest and moral. The average cosine similarity was 0.12 for the provided interest and 0.27 for the moral, demonstrating a balance between incorporating the reader's interest and maintaining the intended moral.\footnote{These cosine similarity values are relatively high; for context, the cosine similarity between the embeddings for \textit{craft} and \textit{carpentry} is 0.164.} A detailed sample of the top terms identified for each story is provided in Appendix Table \ref{tab:BERTopic}.

\begin{table}[t]
\centering
\setlength{\tabcolsep}{8pt} % Adjusted horizontal padding between columns for compactness
\renewcommand{\arraystretch}{1.1} % Adjusted row padding
\small % Reduce font size to fit within a single column

\begin{tabular}{p{4cm}  p{2.5cm}} 
\toprule
\textbf{Story Type} & \textbf{Correctly Identified Morals} \\
\midrule
Generic Human-Written & 23/24 \\
Generic LLM-Generated & 23/23 \\
Personalized LLM-Generated & 25/25 \\
\bottomrule
\end{tabular}
\vspace{-1mm}
\caption{Number of correctly identified morals for each story type, excluding `N/A' responses}
\label{tab:moral}
\end{table}

\begin{figure}[t]
\centering
\begin{tikzpicture}
\begin{axis}[
    ymin=2, ymax=5, 
    ylabel={Average Rating},
    ylabel style={font=\fontsize{10pt}{12pt}\selectfont},
    xlabel style={font=\fontsize{8pt}{10pt}\selectfont, font=\small},
    xticklabel style={font=\footnotesize},
    xtick=data,
    xticklabels={
        Satisfaction, Quality, Engagement, Personalness
    },
    x tick label style={rotate=25,anchor=east},
    legend style={at={(0.5,-0.30)},anchor=north, legend columns=2, font=\scriptsize},
    ymajorgrids=true,
    grid style=dashed,
    cycle list name=color list,
    height=5cm, % Adjusted for double column
    width=7.5cm, % Adjusted for double column
]

% Data for Human Evaluators - Using solid lines
\addplot+[line width=2.5pt, mark=*, mark options={Blue}, draw=Blue, solid] plot coordinates {
    (1, 4.27) (2, 4.19) (3, 4.39) (4, 4.04)
};
\addlegendentry{Human: Pers. LLM}

\addplot+[line width=2.5pt, mark=*, mark options={orange}, draw=orange, solid] plot coordinates {
    (1,3.65) (2,3.81) (3,3.5) (4, 2.62)
};
\addlegendentry{Human: Gen. LLM}

\addplot+[line width=2.5pt, mark=*, mark options={Green}, draw=Green, solid] plot coordinates {
    (1,3.85) (2, 3.81) (3, 3.31) (4,2.5)
};
\addlegendentry{Human: Gen. Human}

% Data for GPT-4 Evaluators - Using dashed lines
\addplot+[line width=2.5pt, mark=square*, mark options={fill=Blue}, draw=Blue, dashed] plot coordinates {
    (1, 4.62) (2, 4.46) (3, 4.5) (4, 4.55)
};
\addlegendentry{GPT-4: Pers. LLM}

\addplot+[line width=2.5pt, mark=square*, mark options={fill=orange}, draw=orange, dashed] plot coordinates {
    (1,3.65) (2,3.55) (3,3.62) (4, 2.15)
};
\addlegendentry{GPT-4: Gen. LLM}

\addplot+[line width=2.5pt, mark=square*, mark options={fill=Green}, draw=Green, dashed] plot coordinates {
    (1,3.65) (2, 3.33) (3, 3.42) (4,2.04)
};
\addlegendentry{GPT-4: Gen. Human}

\end{axis}
\end{tikzpicture}
\caption{Comparative evaluation of narrative types by human and GPT-4 evaluators across different metrics}
\label{fig:combined_results}
\end{figure}
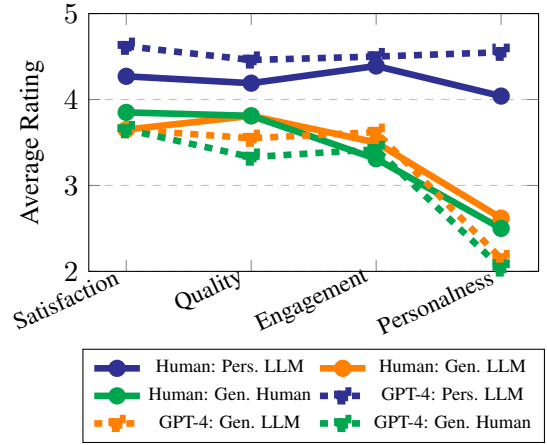

\paragraph{Are \corpusname{} preferred?}
Figure~\ref{fig:combined_results} shows that personalized LLM-generated stories in \corpusname{} are consistently rated higher across all metrics compared to both generic LLM-generated and human-written narratives. This preference is pronounced in evaluations by both humans and GPT-4, with personalized narratives outperforming generic versions, particularly in terms of personalness and engagement where the ratings significantly diverge. 

\begin{table}[t]
\centering
\setlength{\tabcolsep}{4pt} % Adjusted horizontal padding between columns for compactness
\renewcommand{\arraystretch}{1.1} % Adjusted row padding
\small % Reduce font size to fit within a single column
\begin{tabular}{l c} 
\toprule
\textbf{Story Type} & \textbf{SDI} \\ 
\midrule
Generic Human-Written                   & 4.13 \\
Generic LLM-Generated (temp = 1)          & 4.42 \\
Generic LLM-Generated (temp = 1.2)      & 4.82\footnotemark\\
Personalized LLM-Generated (all elements)  & \textbf{4.71} \\
Personalized LLM-Generated (element: `interest') & 4.59 \\
\bottomrule
\end{tabular}
\vspace{-1mm}
\caption{The Shannon Diversity Index (SDI) values for all story types. Values are statistically significant (p < 0.01), as determined by a one-way ANOVA.}
\label{tab:sdi_story_types}
\end{table}
% \footnotetext{At this temperature, stories began to lose coherence.}
\footnotetext{Increasing temperature made the stories lose coherence.}

\paragraph{How does personalization affect  moral comprehension?}
We analyzed the impact of personalization on moral comprehension in stories. Evaluators were asked to identify the main message of each type of story, or provide `N/A' if they could not. We manually assessed the evaluators' responses to the intended morals. Excluding `N/A' responses, the correctly identified morals are detailed in Table \ref{tab:moral}. The results indicate that differences in moral identification across story types are not statistically significant, demonstrating that adding personalization did not negatively affect the model's ability to convey the intended moral. A sample of evaluator responses is shown in Appendix Figure \ref{tab:example_moral}. 

\paragraph{What is the impact of personalization on textual diversity?}
We analyzed how personalization elements impact textual diversity using the Shannon Diversity Index (SDI). Table \ref{tab:sdi_story_types} shows that personalized stories achieve the highest SDI among all story types. Including a single personalization element, such as the `interest' element, also increases SDI compared to generic and human-written stories with the same moral.  Additionally, we observed that increasing GPT-4's temperature negatively affects the diversity and coherence of generic LLM-generated stories. At a temperature of 1.2, the stories showed increased diversity but began to lose coherence. Further increasing the temperature to 1.5 resulted in nonsensical outputs.

%\paragraph{How well do personalized stories integrate identity elements?}
%To assess how well personalized stories align with the provided moral and interests, we used BERTopic for topic modeling to identify the top five terms for each story. We then calculated cosine similarity using Word2Vec embeddings (word2vec-google-news-300) \citep{NIPS2013_9aa42b31} between these top terms and the provided interest and moral. The average cosine similarity between the top 5 terms and the provided interest was 0.12, and 0.27 with the moral, demonstrating a balance between incorporating the reader's interest and maintaining the intended moral. A detailed sample of the top terms identified for each story is provided in Appendix Table \ref{tab:BERTopic}.

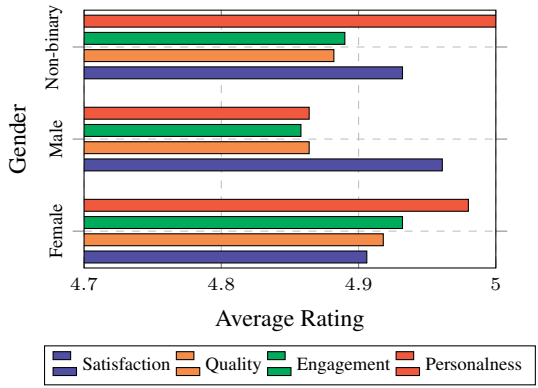
\begin{figure}[t]
\centering
\begin{tikzpicture}
\begin{axis}[
    ylabel={Gender},
    xlabel={Average Rating},
    xlabel style={font=\fontsize{8pt}{10pt}\small},
    ylabel style={font=\fontsize{8pt}{10pt}\small},
    xbar,
    bar width=4.5pt,
    enlarge y limits=0.20,
    ytick=data,
    yticklabels={Female, Male, Non-binary},
    y tick label style={rotate=90},
    yticklabel style={font=\scriptsize}, 
    xticklabel style={font=\scriptsize}, 
    legend style={at={(0.5,-0.3)},anchor=north, font=\scriptsize, legend columns=-1},
    grid=major, % Add major grid lines
    grid style=dashed,
    xmin=4.7, xmax=5.0, 
    width=7cm,
    height=5cm
]

\addplot+[xbar, fill=Blue!80, draw=black] coordinates {
    (4.906,1) (4.961,2) (4.932,3)
};
\addlegendentry{Satisfaction}

\addplot+[xbar, fill=Orange!90, draw=black] coordinates {
    (4.918,1) (4.864,2) (4.882,3)
};
\addlegendentry{Quality}

\addplot+[xbar, fill=Green!90, draw=black] coordinates {
    (4.932,1) (4.858,2) (4.89,3)
};
\addlegendentry{Engagement}

\addplot+[xbar, fill=Red!80, draw=black] coordinates {
    (4.98,1) (4.864,2) (5.000,3)
};
\addlegendentry{Personalness}

\end{axis}
\end{tikzpicture}
\caption{Average ratings by GPT-4 across gender}
\label{fig:evaluation_gender_main}
\vspace{-1mm}
\end{figure}

\paragraph{Are there biases in LLM evaluations of personalized stories?}
We found several preferential biases in GPT-4's evaluation results. Figure \ref{fig:evaluation_gender_main} shows an instance of gender-based bias, with stories featuring non-binary characters receiving the highest personalness ratings, while those with male characters rated lower in quality and engagement. Ethnic background also influences evaluations, with Norwegian and Japanese characters rated higher across all metrics (Appendix Figure \ref{fig:evaluation_ethnicity}). We also observed inter-model preferential biases across the three models used for generating personalized stories, with Claude-3 consistently receiving higher ratings compared to GPT-4 and Gemini-1.5. An overview of all bias results is provided in Appendix \ref{appendix_A3_2}.

\section{Extended Analyses}

\paragraph{Qualitative comparison of human and LLM evaluations}
We examine cases where human and LLM evaluators either contradict or agree on the scores assigned to stories, providing insights into the differences in evaluations and preferences for various types of stories. Examples of these cases, highlighting instances of both agreement and disagreement between human and LLM evaluators, are presented in Appendix Figure \ref{fig:qualitative_comparison}.

\paragraph{Image generation for personalized stories}
We explored the potential of incorporating images into stories to enhance engagement and representation. The image generation and evaluation processes are detailed in Appendix Table \ref{fig:personalization_test_images}. Notably, human evaluators show a high accuracy in identifying personalized elements in the images generated by DALL·E 2 \cite{ramesh2022hierarchical}, with gender and interest being recognized with 100\% and 95\% accuracy, respectively (Appendix Figure \ref{fig:personalization_test_images_results}).

\paragraph{Correlation between human and LLM evaluators}
Correlation analysis revealed a low to moderate alignment between human evaluators and GPT-4 in story evaluation metrics. GPT-4 aligned more closely with human evaluators on quality across all story types (correlations 0.22-0.47), but showed the weakest correlation in assessing personalness, particularly for personalized stories (as low as 0.08). This suggests that while GPT-4 is increasingly used for various evaluation tasks, its effectiveness in assessing subjective aspects of creative tasks is limited. A detailed analysis of these correlations and temperature variations is presented in Appendix \ref{appendix_A31}, Table~\ref{tab:spearman_correlation_results} and Figure~\ref{fig:spearman_temperature}.

\section{Related Work}

Our study builds on research on the effectiveness of personalized narratives in engaging readers and improving learning outcomes \cite{zhang2024mathemyths, 774fe29d-5543-30c3-a709-5dc9035f390b, articlePersonaLLMHirsh}. We extend this work by examining how LLMs can generate personalized narratives to increase reader engagement and satisfaction. While promising, the accuracy of personal traits in generated content remains challenging, with studies showing mixed results \cite{jiang2024personallm, bhandari2023trustworthiness}. Concurrently, LLM exploration in narrative generation has focused on improving coherence and depth \cite{andreas2022language, shen2023storygptv, el-refai2024swag, gomez2023confederacy}. To assess these advancements, recent evaluative techniques for narrative systems emphasize user interactions and alignment metrics between visual content and narratives \cite{el-refai2024swag, ning2023album}.

% \newpage
\section{Conclusion}
Our study demonstrates the potential of LLMs in generating personalized narratives that effectively incorporate diverse identity elements and enhance reader engagement compared to generic stories. \corpusname{} consists of 1,500 personalized stories that consistently outperform generic ones on key metrics. By making \corpusname{} publicly available and integrating it into an interactive web application, we aim to encourage further research on personalized narrative generation, contributing to more engaging and inclusive content. Future work could explore out-group perceptions of these narratives, broadening our understanding of personalization's impact across diverse audiences.

\section*{Limitations}

\paragraph{Story Constraints:} To maintain consistency and feasibility within the scope of our study, we imposed certain constraints on the stories generated, such as limiting the length to 250-300 words and focusing on a specific set of morals. While these constraints allowed for a controlled comparison between personalized and generic stories, they may not fully capture the potential of LLMs in generating longer, more complex narratives or exploring a wider range of themes and morals. Future research could investigate the impact of personalization on stories of varying lengths and themes to gain a more comprehensive understanding of how these factors influence reader engagement and satisfaction.

\paragraph{Demographic Diversity:} While our study aimed to include a diverse range of identities and backgrounds, the demographic diversity of our human evaluators was by no means the perfect sample of global readership. The majority of our evaluators were university students, which may not be representative of the broader population. Future research should include a more diverse pool of evaluators across age, education, and cultural backgrounds to ensure the generalizability of the findings and to capture a wider range of perspectives on personalized storytelling.

% \vspace{10pt}

\paragraph{Scope of Personalization:} Our study primarily examined personalization factors like age, gender, interests, and ethnic background. However, aspects such as personality traits, emotional resonance, and narrative preferences were not extensively investigated but could notably enhance engagement and narrative impact. For example, aligning story elements with reader emotional responses or tailoring narratives to specific preferences like mystery, romance, or adventure could significantly boost satisfaction and engagement.

% \vspace{8pt}

\paragraph{Subjectivity of Evaluation:} Another limitation of our study is the inherent subjectivity involved in evaluating the impact of personalized stories. Despite our attempts to standardize evaluation criteria and maintain consistency among evaluators, individual preferences, biases, and interpretations can still significantly influence the outcomes. This subjectivity can lead to variability in how different evaluators perceive and rate the same narrative elements.

\paragraph{Model Selection and Variety:} Our study utilized GPT-4, Claude3, Gemini-1.5, and DALL·E 2  for generating and evaluating narratives and images. This limited selection may affect the generalizability of our findings, as different models might produce or assess stories differently based on their training data and algorithms. Expanding future research to include a variety of models, including open-source ones, could provide a more comprehensive understanding of how different language models handle personalization in storytelling and evaluate narrative elements.

\section*{Ethical Considerations}

We followed strict ethical standards throughout our research to ensure validity and fairness. Consent and transparency were central to our approach, with all participants fully informed and providing explicit consent. We also ensured compliance with intellectual property rights by using Aesop's fables, which are in the public domain.

\paragraph{Data Privacy and Security:} Ensuring the privacy and security of participants' personal information was a top priority. We collected and used personal details such as age, gender, interests, and ethnic background to generate personalized stories. Robust data protection measures were implemented, including secure storage, anonymization, and restricted access to sensitive information. Participants were informed about how their data would be used, stored, and protected.

\paragraph{Potential Misuse and Unintended Consequences:} While personalized storytelling has the potential to enhance engagement and representation, we carefully considered the potential for misuse or unintended consequences. To mitigate risks such as the manipulation of individuals' emotions or the reinforcement of stereotypes, we implemented safeguards against harmful content and regularly audited the generated stories for potential biases or inappropriate themes.

\paragraph{Inclusivity and Representation:} When generating personalized stories, we strived to ensure that the stories were inclusive and representative of diverse identities and experiences. This included considering factors such as race, ethnicity, gender identity, sexual orientation, disability, and socioeconomic status. We aimed to create stories that were respectful, authentic, and empowering for all individuals, avoiding stereotypes and promoting positive representation.

Accountability and integrity were paramount in reporting our results, including limitations and implications. Additionally, every narrative generated by LLMs underwent a thorough review to maintain quality and appropriateness, enhancing the reliability of our findings and participant well-being.

\section*{Acknowledgements}
This work was supported by the Natural Sciences and Engineering Research Council of Canada and by the New Frontiers in Research Fund.

\bibliography{custom}

\begin{thebibliography}{37}
\expandafter\ifx\csname natexlab\endcsname\relax\def\natexlab#1{#1}\fi

\bibitem[{Andreas(2022)}]{andreas2022language}
Jacob Andreas. 2022.
\newblock \href {http://arxiv.org/abs/2212.01681} {Language models as agent models}.

\bibitem[{Babatunde et~al.(2024)Babatunde, Odejide, Edunjobi, and Ogundipe}]{article2}
Sodiq Babatunde, Opeyemi Odejide, Tolulope Edunjobi, and Damilola Ogundipe. 2024.
\newblock \href {https://doi.org/10.51594/ijmer.v6i3.964} {The role of ai in marketing personalization: A theoretical exploration of consumer engagement strategies}.
\newblock \emph{International Journal of Management \& Entrepreneurship Research}, 6:936--949.

\bibitem[{Bhandari and Brennan(2023)}]{bhandari2023trustworthiness}
Prabin Bhandari and Hannah~Marie Brennan. 2023.
\newblock Trustworthiness of children stories generated by large language models.
\newblock \emph{arXiv preprint arXiv:2308.00073}.

\bibitem[{Bishop(1990)}]{bishop1990mirrors}
Rudine~Sims Bishop. 1990.
\newblock Mirrors, windows, and sliding glass doors. perspectives: Choosing and using books for the classroom, 6 (3).
\newblock \emph{Perspectives: Choosing and using books for the classroom}, 6(3):ix--xi.

\bibitem[{Brown et~al.(2020)Brown, Mann, Ryder, Subbiah, Kaplan, Dhariwal, Neelakantan, Shyam, Sastry, Askell et~al.}]{brown2020language}
Tom Brown, Benjamin Mann, Nick Ryder, Melanie Subbiah, Jared~D Kaplan, Prafulla Dhariwal, Arvind Neelakantan, Pranav Shyam, Girish Sastry, Amanda Askell, et~al. 2020.
\newblock Language models are few-shot learners.
\newblock \emph{Advances in neural information processing systems}, 33:1877--1901.

\bibitem[{{CCBC}(2021)}]{ccbc2021}
{CCBC}. 2021.
\newblock \href {https://ccbc.education.wisc.edu/literature-resources/ccbc-diversity-statistics/books-by-about-poc-fnn/} {Books by and/or about black, indigenous, and people of color (all years)}.
\newblock Data retrieved from the Cooperative Children's Book Center.

\bibitem[{Chowdhery et~al.(2022)Chowdhery, Narang, Devlin, Bosma, Mishra, Roberts, Barham, Chung, Sutton, Gehrmann et~al.}]{chowdhery2022palm}
Aakanksha Chowdhery, Sharan Narang, Jacob Devlin, Maarten Bosma, Gaurav Mishra, Adam Roberts, Paul Barham, Hyung~Won Chung, Charles Sutton, Sebastian Gehrmann, et~al. 2022.
\newblock Palm: Scaling language modeling with pathways.
\newblock \emph{arXiv preprint arXiv:2204.02311}.

\bibitem[{El-Refai et~al.(2024)El-Refai, Patel, and Pei}]{el-refai2024swag}
Karim El-Refai, Zeeshan Patel, and Jonathan Pei. 2024.
\newblock Swag: Storytelling with action guidance.
\newblock \emph{arXiv preprint arXiv:2402.03483}.

\bibitem[{Fleming et~al.(2016)Fleming, Catapano, Thompson, and Carrillo}]{fleming2016more}
Jane Fleming, Susan Catapano, Candace~M Thompson, and Sandy~Ruvalcaba Carrillo. 2016.
\newblock \emph{More mirrors in the classroom: Using urban children's literature to increase literacy}.
\newblock Rowman \& Littlefield.

\bibitem[{Galitsky(2024)}]{202402.1709}
Boris~A. Galitsky. 2024.
\newblock \href {https://doi.org/10.20944/preprints202402.1709.v1} {Llm- based personalized recommendations in health}.
\newblock \emph{Preprints}.

\bibitem[{Gilardi et~al.(2023)Gilardi, Alizadeh, and Kubli}]{Gilardi_2023}
Fabrizio Gilardi, Meysam Alizadeh, and Maël Kubli. 2023.
\newblock \href {https://doi.org/10.1073/pnas.2305016120} {Chatgpt outperforms crowd workers for text-annotation tasks}.
\newblock \emph{Proceedings of the National Academy of Sciences}, 120(30).

\bibitem[{G{\'o}mez-Rodr{\'i}guez and Williams(2023)}]{gomez2023confederacy}
Carlos G{\'o}mez-Rodr{\'i}guez and Paul Williams. 2023.
\newblock A confederacy of models: a comprehensive evaluation of llms on creative writing.
\newblock In \emph{Findings of the Association for Computational Linguistics: EMNLP 2023}, pages 14504--14528, A Coru{\~n}a, Spain and Sunshine Coast, Australia. Association for Computational Linguistics.

\bibitem[{Grootendorst(2022)}]{grootendorst2022bertopic}
Maarten Grootendorst. 2022.
\newblock Bertopic: Neural topic modeling with a class-based tf-idf procedure.
\newblock \emph{arXiv preprint arXiv:2203.05794}.

\bibitem[{Heineke et~al.(2022)Heineke, Papola-Ellis, and Elliott}]{https://doi.org/10.1002/trtr.2139}
Amy~J. Heineke, Aimee Papola-Ellis, and Joseph Elliott. 2022.
\newblock \href {https://doi.org/https://doi.org/10.1002/trtr.2139} {Using texts as mirrors: The power of readers seeing themselves}.
\newblock \emph{The Reading Teacher}, 76(3):277--284.

\bibitem[{Hirsh and Peterson(2009)}]{articlePersonaLLMHirsh}
Jacob Hirsh and Jordan Peterson. 2009.
\newblock \href {https://doi.org/10.1016/j.jrp.2009.01.006} {Personality and language use in self-narratives}.
\newblock \emph{Journal of Research in Personality}, 43:524--527.

\bibitem[{Hoytt et~al.(2022)Hoytt, Hunt, and Lovett}]{hoytt2022impact}
Karima Hoytt, Sherrica Hunt, and Margaret~A Lovett. 2022.
\newblock Impact of cultural responsiveness on student achievement in secondary schools.
\newblock \emph{Alabama Journal of Educational Leadership}, 9:1--12.

\bibitem[{Huyck and Dahlen(2019)}]{huyck2019diversity}
D~Huyck and SP~Dahlen. 2019.
\newblock Diversity in children’s books 2018. sarahpark. com blog. created in consultation with edith campbell, molly beth griffin, kt horning, debbie reese, ebony elizabeth thomas, and madeline tyner, with statistics compiled by the cooperative children’s book center, school of education, university of wisconsin-madison.

\bibitem[{Jiang et~al.(2023)Jiang, Xu, Zhu, Han, Zhang, and Zhu}]{jiang2023evaluating}
Guangyuan Jiang, Manjie Xu, Song-Chun Zhu, Wenjuan Han, Chi Zhang, and Yixin Zhu. 2023.
\newblock \href {http://arxiv.org/abs/2206.07550} {Evaluating and inducing personality in pre-trained language models}.

\bibitem[{Jiang et~al.(2024)Jiang, Zhang, Cao, Breazeal, Roy, and Kabbara}]{jiang2024personallm}
Hang Jiang, Xiajie Zhang, Xubo Cao, Cynthia Breazeal, Deb Roy, and Jad Kabbara. 2024.
\newblock \href {http://arxiv.org/abs/2305.02547} {Personallm: Investigating the ability of large language models to express personality traits}.

\bibitem[{Li et~al.(2024)Li, Dou, Lv, Liu, Xu, Wu, Ling, Zheng, and Huang}]{li2024tailoring}
Tianlong Li, Shihan Dou, Changze Lv, Wenhao Liu, Jianhan Xu, Muling Wu, Zixuan Ling, Xiaoqing Zheng, and Xuanjing Huang. 2024.
\newblock \href {http://arxiv.org/abs/2310.16582} {Tailoring personality traits in large language models via unsupervisedly-built personalized lexicons}.

\bibitem[{Malik et~al.(2024)Malik, Bernard, Pauchet, Chatelain, Picot-Clémente, and Cortinovis}]{10400468}
Usman Malik, Simon Bernard, Alexandre Pauchet, Clément Chatelain, Romain Picot-Clémente, and Jérôme Cortinovis. 2024.
\newblock \href {https://doi.org/10.1109/ACCESS.2024.3354705} {Pseudo-labeling with large language models for multi-label emotion classification of french tweets}.
\newblock \emph{IEEE Access}, 12:15902--15916.

\bibitem[{Mao et~al.(2024)Mao, Wang, Wang, Jiang, Xie, Huang, and Zhang}]{mao2024editing}
Shengyu Mao, Xiaohan Wang, Mengru Wang, Yong Jiang, Pengjun Xie, Fei Huang, and Ningyu Zhang. 2024.
\newblock \href {http://arxiv.org/abs/2310.02168} {Editing personality for large language models}.

\bibitem[{Mikolov et~al.(2013)Mikolov, Sutskever, Chen, Corrado, and Dean}]{NIPS2013_9aa42b31}
Tomas Mikolov, Ilya Sutskever, Kai Chen, Greg~S Corrado, and Jeff Dean. 2013.
\newblock \href {https://proceedings.neurips.cc/paper_files/paper/2013/file/9aa42b31882ec039965f3c4923ce901b-Paper.pdf} {Distributed representations of words and phrases and their compositionality}.
\newblock In \emph{Advances in Neural Information Processing Systems}, volume~26. Curran Associates, Inc.

\bibitem[{Ning et~al.(2023)Ning, Xie, Chen, Song, Yuan, Tian, and Ye}]{ning2023album}
Munan Ning, Yujia Xie, Dongdong Chen, Zeyin Song, Lu~Yuan, Yonghong Tian, and Qixiang Ye. 2023.
\newblock Album storytelling with iterative story-aware captioning and large language models.
\newblock \emph{arXiv preprint arXiv:2305.12943}.

\bibitem[{OpenAI(2023)}]{openai2023gpt4}
OpenAI. 2023.
\newblock \href {http://arxiv.org/abs/2303.08774} {Gpt-4 technical report}.

\bibitem[{Pennebaker and Graybeal(2001)}]{774fe29d-5543-30c3-a709-5dc9035f390b}
James~W. Pennebaker and Anna Graybeal. 2001.
\newblock \href {http://www.jstor.org/stable/20182707} {Patterns of natural language use: Disclosure, personality, and social integration}.
\newblock \emph{Current Directions in Psychological Science}, 10(3):90--93.

\bibitem[{Phillips(2014)}]{article_phil}
Katherine Phillips. 2014.
\newblock \href {https://doi.org/10.1038/scientificamerican1014-42} {How diversity works}.
\newblock \emph{Scientific American}, 311:42--7.

\bibitem[{Ramesh et~al.(2022)Ramesh, Dhariwal, Nichol, Chu, and Chen}]{ramesh2022hierarchical}
Aditya Ramesh, Prafulla Dhariwal, Alex Nichol, Casey Chu, and Mark Chen. 2022.
\newblock \href {http://arxiv.org/abs/2204.06125} {Hierarchical text-conditional image generation with clip latents}.

\bibitem[{Reid et~al.(2024)Reid, Savinov, Teplyashin, Lepikhin, Lillicrap, Alayrac, Soricut, Lazaridou, Firat, Schrittwieser et~al.}]{reid2024gemini}
Machel Reid, Nikolay Savinov, Denis Teplyashin, Dmitry Lepikhin, Timothy Lillicrap, Jean-baptiste Alayrac, Radu Soricut, Angeliki Lazaridou, Orhan Firat, Julian Schrittwieser, et~al. 2024.
\newblock Gemini 1.5: Unlocking multimodal understanding across millions of tokens of context.
\newblock \emph{arXiv preprint arXiv:2403.05530}.

\bibitem[{Shen and Elhoseiny(2023)}]{shen2023storygptv}
Xiaoqian Shen and Mohamed Elhoseiny. 2023.
\newblock Storygpt-v: Large language models as consistent story visualizers.
\newblock \emph{arXiv preprint arXiv:2402.03483}.

\bibitem[{Tarkka et~al.(2024)Tarkka, Koljonen, Korhonen, Laine, Martiskainen, Elo, and Laippala}]{tarkka-etal-2024-automated}
Otto Tarkka, Jaakko Koljonen, Markus Korhonen, Juuso Laine, Kristian Martiskainen, Kimmo Elo, and Veronika Laippala. 2024.
\newblock \href {https://aclanthology.org/2024.parlaclarin-1.11} {Automated emotion annotation of {F}innish parliamentary speeches using {GPT}-4}.
\newblock In \emph{Proceedings of the IV Workshop on Creating, Analysing, and Increasing Accessibility of Parliamentary Corpora (ParlaCLARIN) @ LREC-COLING 2024}, pages 70--76, Torino, Italia. ELRA and ICCL.

\bibitem[{Touvron et~al.(2023)Touvron, Lavril, Izacard, Martinet, Lachaux, Lacroix, Rozi{\`e}re, Goyal, Hambro, Azhar et~al.}]{touvron2023llama}
Hugo Touvron, Thibaut Lavril, Gautier Izacard, Xavier Martinet, Marie-Anne Lachaux, Timoth{\'e}e Lacroix, Baptiste Rozi{\`e}re, Naman Goyal, Eric Hambro, Faisal Azhar, et~al. 2023.
\newblock Llama: Open and efficient foundation language models.
\newblock \emph{arXiv preprint arXiv:2302.13971}.

\bibitem[{Walkington and Bernacki(2014)}]{inbook}
Candace Walkington and Matthew Bernacki. 2014.
\newblock \href {https://doi.org/10.1108/S0749-742320140000018004} {\emph{Motivating Students by “Personalizing” Learning around Individual Interests: A Consideration of Theory, Design, and Implementation Issues}}, volume~18, chapter~4. Preprints.

\bibitem[{Wang et~al.(2024)Wang, Xiao, tse Huang, Yuan, Xu, Guo, Tu, Fei, Leng, Wang, Chen, Li, and Xiao}]{wang2024incharacter}
Xintao Wang, Yunze Xiao, Jen tse Huang, Siyu Yuan, Rui Xu, Haoran Guo, Quan Tu, Yaying Fei, Ziang Leng, Wei Wang, Jiangjie Chen, Cheng Li, and Yanghua Xiao. 2024.
\newblock \href {http://arxiv.org/abs/2310.17976} {Incharacter: Evaluating personality fidelity in role-playing agents through psychological interviews}.

\bibitem[{Wier et~al.(1890)Wier, Tenniel, and Griset}]{wier1890aesop}
H.~Wier, J.~Tenniel, and E.H. Griset. 1890.
\newblock \href {https://books.google.ca/books?id=UGcSqAAACAAJ} {\emph{Aesop's Fables: A New Revised Version from Original Sources}}.
\newblock Worthington, Company.

\bibitem[{Zhang et~al.(2024)Zhang, Liu, Ziska, Jeon, Yu, and Xu}]{zhang2024mathemyths}
Chao Zhang, Xuechen Liu, Katherine Ziska, Soobin Jeon, Chi-Lin Yu, and Ying Xu. 2024.
\newblock \href {https://doi.org/10.1145/3613904.3642647} {Mathemyths: Leveraging large language models to teach mathematical language through child-ai co-creative storytelling}.
\newblock In \emph{Proceedings of the CHI Conference on Human Factors in Computing Systems}, pages 1--23, New York, NY, USA. ACM.

\bibitem[{Zhao et~al.(2023)Zhao, Zhou, Li, Tang, Wang, Hou, Min, Zhang, Zhang, Dong, Du, Yang, Chen, Chen, Jiang, Ren, Li, Tang, Liu, Liu, Nie, and Wen}]{zhao2023survey}
Wayne~Xin Zhao, Kun Zhou, Junyi Li, Tianyi Tang, Xiaolei Wang, Yupeng Hou, Yingqian Min, Beichen Zhang, Junjie Zhang, Zican Dong, Yifan Du, Chen Yang, Yushuo Chen, Zhipeng Chen, Jinhao Jiang, Ruiyang Ren, Yifan Li, Xinyu Tang, Zikang Liu, Peiyu Liu, Jian-Yun Nie, and Ji-Rong Wen. 2023.
\newblock \href {http://arxiv.org/abs/2303.18223} {A survey of large language models}.

\end{thebibliography}

\clearpage
\onecolumn

\appendix
\section{Appendix}
\label{sec:appendix}

\begin{figure*}[h]
    \centering
    \includegraphics[width=11.2cm]{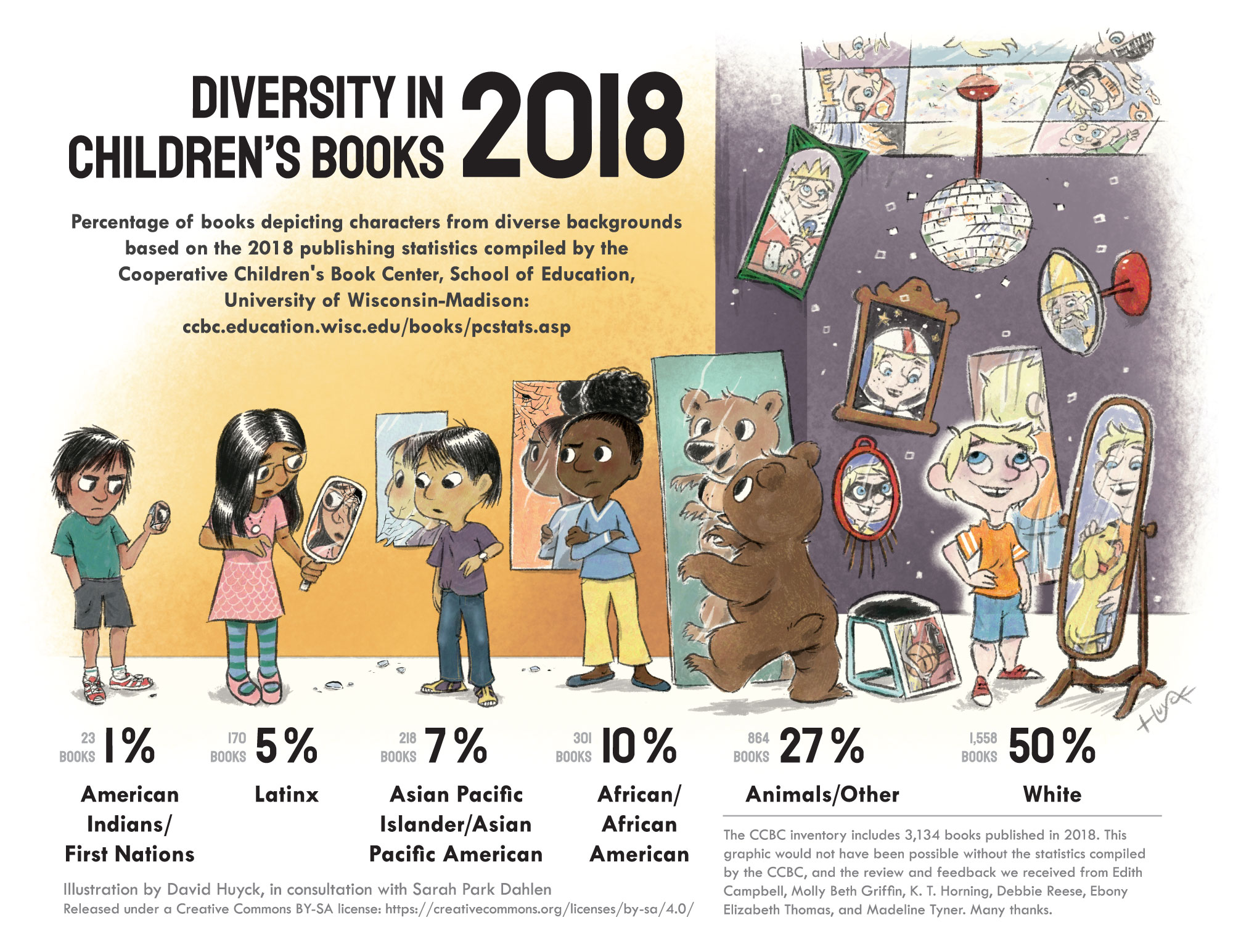}

\caption{Illustration of diversity representation in children's books based on 2018 publishing statistics. Data derived from the Cooperative Children's Book Center, University of Wisconsin-Madison. (Illustration by David Huyck, in consultation with Sarah Park Dahlen) \cite{huyck2019diversity}.}
    \label{fig:diversity}
\end{figure*}

\begin{figure}[h]
% \centering
% \begin{figure*}[ht]/
\begin{tcolorbox}[sharp corners, boxrule=0.2mm, colback=white, 
title={GPT-4 as an Evaluator}, 
coltitle=white, colbacktitle=darkgray, fonttitle=\bfseries\centering]

% \begin{tcolorbox}[sharp corners, boxrule=0.2mm, colback=white]

\small \textbf{Sample GPT-4 INPUT:}

\small You are ANDRII, a MALE character, aged 25, of RUSSIAN ethnicity, with an interest in WRESTLING. As an evaluator, evaluate the story based on the following metrics:

\vspace{0.1cm}
METRICS:
\vspace{0.1cm}
    \\ SATISFACTION: How would you rate your overall satisfaction with the story? 
        \\ \small 1 (BAD): The story did not meet any expectations, had significant flaws in plot, character development, or other key areas, and provided little to no enjoyment or value.
        \\ \small 5 (GOOD): The story exceeded all expectations, was exceptional in most if not all aspects.
    \\ QUALITY: How would you rate the quality of the story? ...
\vspace{0.15cm}
\\
STORY:

\small In the heart of Moscow lived Andrii, a 25-year-old whose heart and soul were intertwined with the art of wrestling. Every morning, before the first ray of sunlight kissed the domes of Saint Basil's Cathedral ...

\vspace{0.2cm}

\textbf{Sample GPT-4 OUTPUT:}

I would rate my overall satisfaction with the story a 5. The unexpected turn in the plot, well-developed character, and the consistent twists kept me engaged, and the story exceeded all my expectations.

I would rate the quality of content of the story a 4. The intertwining of fiction and reality, the character ...

\end{tcolorbox}
\caption{Sample input and output of GPT-4 evaluating a personalized story}
\label{GPT4_Evaluator}
\end{figure}

% \captionof{figure}{A sample of input and output of GPT-4 story evaluation process. }
% \label{GPT4_Evaluator}
% % \end{figure*}

\begin{table*}[h]
    \centering
    \small
    \begin{tabularx}{\textwidth}{|p{2cm}|p{3.7cm}|p{5cm}|X|} 
        \toprule
       \textbf{Annotator \#} &\textbf{Story Type} &Q: \textbf{``What is the main message or moral of the story? If you can't identify, write `N/A' `'}.& \textbf{Actual Moral of the Story }\\ 
\midrule
Annotator \#1 & Generic Human-Written & To preserver and never quit. The tortoise never stopped during the race but the hare did and ended up losing because of it. & The race is not always to the swift.
\\
\midrule
Annotator \#1 &Generic LLM-Generated&Perseverance. Keep your eye on the prize, don’t worry about what others are doing.  &The race is not always to the swift.
\\ 
\midrule
Annotator \#1  &Personalized LLM-Generated&Perseverance. Not everyone’s goals in life will be the same and will take time. Keep taking steps towards your goal and you will be rewarded. &The race is not always to the swift.
\\
\midrule
Annotator \#2 & Generic Human-Written & Despite one's status someone can still help another. & Kindness is never wasted.
\\
\midrule
Annotator \#2 &Generic LLM-Generated&DIn order to truly help somebody, one should do their best to accommodate their needs and avoid demanding some kind of compensation for their troubles.&Kindness is never wasted.
\\ 
\midrule
Annotator \#2  &Personalized LLM-Generated&That kindness has a way of repaying itself one day.&Kindness is never wasted.
\\
    \bottomrule
    \end{tabularx}
\caption{Sample responses from two annotators on the main message or moral of each story type, compared with the actual intended moral of the stories}

    \label{tab:example_moral}
\end{table*}

\newpage
\subsection{Annotators and Dataset Diversity}
\label{sec:appendix_A1}

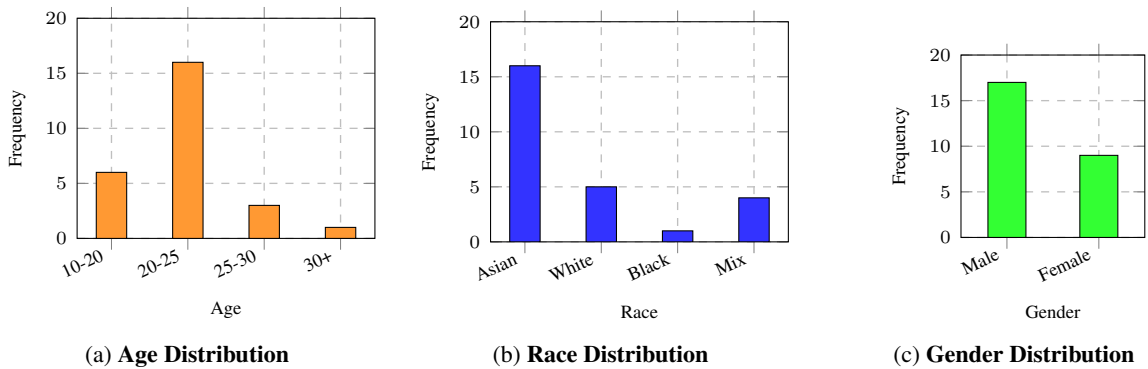
\begin{figure}[H]
    \centering
    \begin{subfigure}{0.32\textwidth}
        \centering
        \begin{tikzpicture}
        \begin{axis}[
            xlabel={Age},
            ylabel={Frequency},
            ylabel style={font=\fontsize{7pt}{11pt}\selectfont},
            xlabel style={font=\fontsize{7pt}{11pt}\selectfont},
            ybar,
            bar width=0.4cm,
            symbolic x coords={10-20, 20-25, 25-30, 30+},
            xtick=data,
            xticklabel style={font=\scriptsize},
            yticklabel style={font=\scriptsize},
            x tick label style={rotate=25,anchor=east},
            grid=major,
            ymin=0, ymax=20,
            width=5.5cm,
            height=4.5cm,
            enlarge x limits=0.15,
            grid style=dashed,
            major grid style={line width=0.5pt,draw=gray!50},
        ]
        \addplot+[ybar, fill=orange!80, draw=black] coordinates {
            (10-20,6) (20-25,16) (25-30,3) (30+,1)
        };
        \end{axis}
        \end{tikzpicture}
        \caption{\small \textbf{Age Distribution}}
    \end{subfigure}
    \hfill
    \begin{subfigure}{0.32\textwidth}
        \centering
        \begin{tikzpicture}
        \begin{axis}[
            xlabel={Race},
            ylabel={Frequency},
            ylabel style={font=\fontsize{7pt}{11pt}\selectfont},
            xlabel style={font=\fontsize{7pt}{11pt}\selectfont},
            ybar,
            bar width=0.4cm,
            symbolic x coords={Asian, White, Black, Mix},
            xtick=data,
            xticklabel style={font=\scriptsize},
            yticklabel style={font=\scriptsize},
            x tick label style={rotate=25,anchor=east},
            grid=major,
            ymin=0, ymax=20,
            width= 5.5cm,
            height=4.5cm,
            enlarge x limits=0.15,
            grid style=dashed,
            major grid style={line width=0.5pt,draw=gray!50},
        ]
        \addplot+[ybar, fill=blue!80, draw=black] coordinates {
            (Asian,16) (White,5) (Black,1) (Mix,4)
        };
        \end{axis}
        \end{tikzpicture}
        \caption{\small \textbf{Race Distribution}}
    \end{subfigure}
    \hfill
    \begin{subfigure}{0.32\textwidth}
        \centering
        \begin{tikzpicture}
        \begin{axis}[
            xlabel={Gender},
            ylabel={Frequency},
            ylabel style={font=\fontsize{7pt}{11pt}\selectfont},
            xlabel style={font=\fontsize{7pt}{11pt}\selectfont},
            ybar,
            bar width=0.5cm,
            symbolic x coords={Male, Female},
            xtick=data,
            xticklabel style={font=\scriptsize},
            yticklabel style={font=\scriptsize},
            x tick label style={rotate=25,anchor=east},
            grid=major,
            ymin=0, ymax=20,
            width=4 cm,
            height=4cm,
            enlarge x limits=0.5,
            grid style=dashed,
            major grid style={line width=0.5pt,draw=gray!50},
        ]
        \addplot+[ybar, fill=green!80, draw=black] coordinates {
            (Male,17) (Female,9)
        };
        \end{axis}
        \end{tikzpicture}
        \caption{\small \textbf{Gender Distribution}}
    \end{subfigure}

    \caption{Demographic distribution of annotators by age, race, and gender}
    \label{fig:annotators_demongraphics}
\end{figure}

% \subsection{Dataset Diversity}
\begin{figure}[!h]

    \centering
    \begin{minipage}[t]{0.45\textwidth}
        \centering
        \begin{tikzpicture}
        \pie[
            text=legend,
            radius=1.8,
            color={Orange!90, Blue!60, Green!70},
            sum=auto,        
            before number=\footnotesize,
            after number=\%,
            text=legend,
            nodes={scale=0.7}
        ]{
            48.3/\small Male,
            47.7/\small Female,
            4.0/\small Non-binary
        }
        \end{tikzpicture}
        \caption{\small Gender Distribution in \corpusname{}}
        \label{fig:gender_mirror}
    \end{minipage}
    % \hfill
    \begin{minipage}[t]{0.45\textwidth}
        \centering
        \begin{tikzpicture}
        \begin{axis}[
            xlabel={Age},
            ylabel={Frequency},
            ylabel style={font=\fontsize{9pt}{11pt}\selectfont},
            xlabel style={font=\fontsize{9pt}{11pt}\selectfont},
            ybar,
            bar width=0.4cm,
            symbolic x coords={10-18, 18-26, 26-35, 35-43, 43-51, 51-60},
            xtick=data,
            xticklabel style={font=\scriptsize},
            yticklabel style={font=\scriptsize},
            x tick label style={rotate=25,anchor=east},
            grid=major,
            ymin=0, ymax=300,
            width=5.5cm,
            height=4.5cm,
            enlarge x limits=0.15,
            grid style=dashed,
            major grid style={line width=0.5pt,draw=gray!50},
        ]
        \addplot+[ybar, fill=Blue!90, draw=black] coordinates {
            (10-18,292) (18-26,235) (26-35,251) (35-43,267) (43-51,209) (51-60,246)
        };
        \end{axis}
        \end{tikzpicture}
        \caption{\small Age Distribution in \corpusname{}}
        \label{fig:age_mirror}
    \end{minipage}
\end{figure}

\subsection{Questionnaire for Annotators}
\label{appendix_A2}

\begin{figure*}[h]
    \centering
    \includegraphics[width=\textwidth]{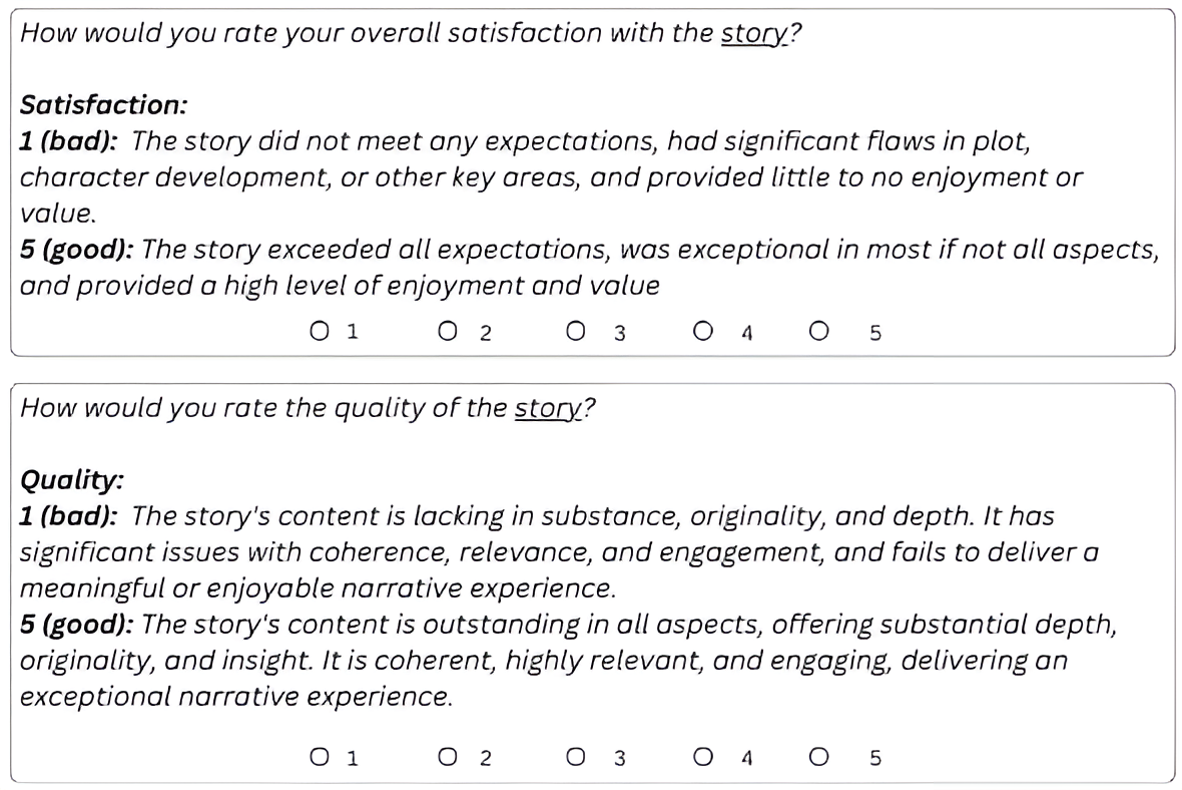}
    % \caption{Satisfaction, Quality, Engagement, and Personalness questions.}
% \label{fig:metrics_questions}
\end{figure*}

\begin{figure*}[!h]
    \centering
    \includegraphics[width=\textwidth]{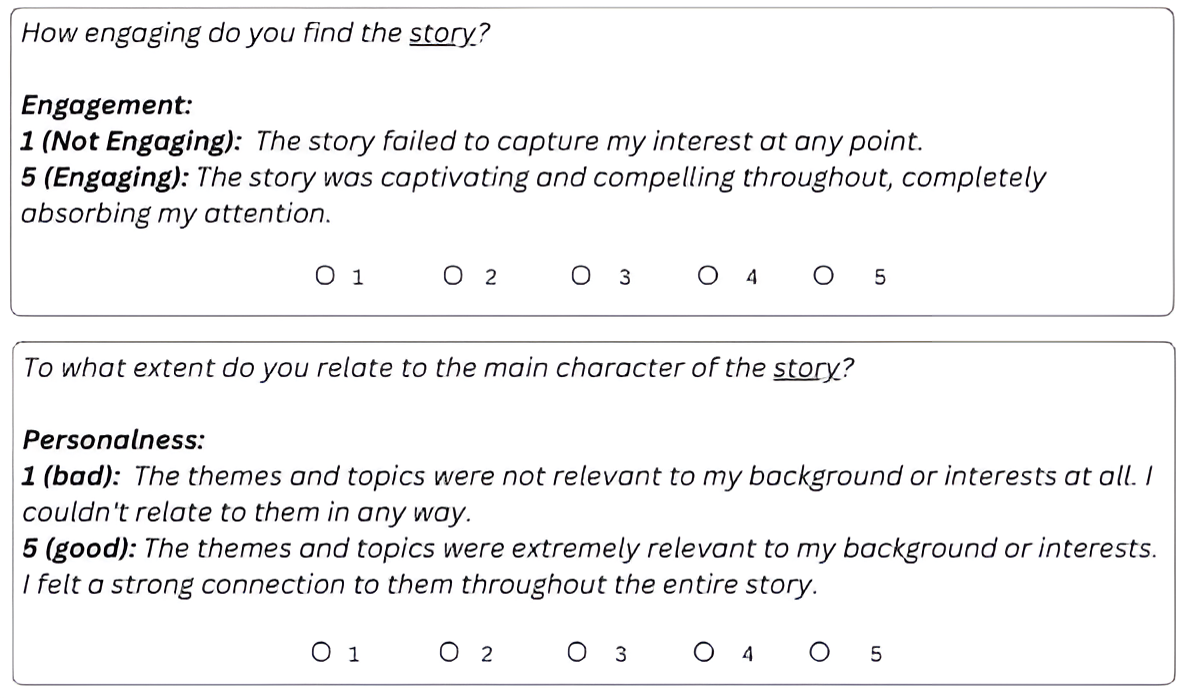}
    \caption{Questionnaire used to assess story satisfaction, quality, engagement, and personalness}
\label{fig:metrics_questions}
\end{figure*}

% \clearpage
\newpage

\subsection{Personalization Example}

\begin{table}[h]
\centering
\setlength{\tabcolsep}{4pt} % Adjusted horizontal padding between columns
\renewcommand{\arraystretch}{1.5} % Adjusted row padding
\scriptsize % Reduce font size to fit within a single column
\resizebox{\textwidth}{!}{ % Resize table to fit within text width
\begin{tabular}{p{2.2cm}|p{6cm}|p{4cm}} % Set specific column widths
\toprule
\textbf{Personalization Elements} & \textbf{Personalized Story} & \textbf{Top 5 Term} \\ 
\midrule
Aveline, 19, Non-binary, Reading, French & 
Aveline was a confluence of distinctive characteristics; their name a symbol of French roots and their gender identity, non-binary, a testament to their unfettered self-expression. At 19, they were a sagacious soul, finding immense joy in ... & 
library (0.033), truth (0.030), guilt (0.023), joy (0.023), lie (0.023) \\

Farida, 23, Female, Carpentry, Uzbek & 
In the heart of the bustling Uzbek city, Tashkent, Farida, a young, passionate woman, was determined to carve out a unique reputation for herself. At 23, she was an anomaly in her city; she was not working a typical job like ... & 
craft (0.0397), reputation (0.039), wood (0.039), client (0.029), city (0.029) \\

Rami, 21, Male, Trekking, Syrian & 
Rami, a 21-year-old Syrian youth, was known in his community for two things — his irresistible passion for trekking and his firm belief in honesty. Dark-haired, with  ... & 
community (0.030), truth (0.028), life (0.022), sun (0.022), adventurous (0.021) \\

\bottomrule
\end{tabular}
}
\caption{This table presents the personalization elements for three individuals, their personalized stories, and the top 5 terms identified by BERTopic for each story, along with the corresponding relevance scores.}

\label{tab:BERTopic}
\end{table}

% \newpage
\subsubsection{Preferential Bias Analysis}
\label{appendix_A3_2}

\begin{figure}[h]
\centering
\begin{tikzpicture}
\begin{axis}[
    ylabel={Model},
    xlabel={Average Rating},
    ylabel style={font=\fontsize{8pt}{10pt}\selectfont},
    xlabel style={font=\fontsize{8pt}{10pt}\selectfont},
    xbar,
    bar width=4.5pt, % Reduced bar width
    enlarge y limits=0.20, % Adjusted for better fit
    ytick=data,
    yticklabels={Gemini-1.5, Claude-3, GPT-4},
    y tick label style={rotate=90},
    yticklabel style={font=\scriptsize}, 
    xticklabel style={font=\scriptsize}, 
    legend style={at={(0.5,-0.3)},anchor=north, font=\small, legend columns=-1},
    grid=major, % Add major grid lines
    grid style=dashed,
    xmin=4.7, xmax=5.0, 
    width=7cm, % Reduced width
    height=4.5cm % Reduced height
]

\addplot+[xbar, fill=Blue!80, draw=black] coordinates {
    (4.812,1) (4.961,2) (4.906,3)
};
\addlegendentry{Satisfaction}

\addplot+[xbar, fill=Orange!90, draw=black] coordinates {
    (4.819,1) (4.962,2) (4.915,3)
};
\addlegendentry{Quality}

\addplot+[xbar, fill=Green!90, draw=black] coordinates {
    (4.829,1) (4.984,2) (4.930,3)
};
\addlegendentry{Engagement}

\addplot+[xbar, fill=Red!80, draw=black] coordinates {
    (4.990,1) (4.996,2) (5.000,3)
};
\addlegendentry{Personalness}

\end{axis}
\end{tikzpicture}
\caption{Average evaluation ratings by GPT-4 across models}
\label{fig:evaluation_model}
\end{figure}
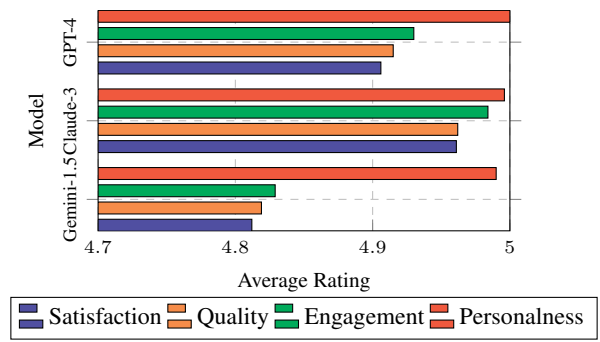

\begin{figure}[h]
\centering
\begin{tikzpicture}
\begin{axis}[
    ylabel={Gender},
    xlabel={Average Rating},
    xlabel style={font=\fontsize{8pt}{10pt}\small},
    ylabel style={font=\fontsize{8pt}{10pt}\small},
    xbar,
    bar width=4.5pt,
    enlarge y limits=0.20,
    ytick=data,
    yticklabels={Female, Male, Non-binary},
    y tick label style={rotate=90},
    yticklabel style={font=\scriptsize}, 
    xticklabel style={font=\scriptsize}, 
    legend style={at={(0.5,-0.3)},anchor=north, font=\small, legend columns=-1},
    grid=major, % Add major grid lines
    grid style=dashed,
    xmin=4.7, xmax=5.0, 
    width=7cm,
    height=4.5cm
]

\addplot+[xbar, fill=Blue!80, draw=black] coordinates {
    (4.906,1) (4.961,2) (4.932,3)
};
\addlegendentry{Satisfaction}

\addplot+[xbar, fill=Orange!90, draw=black] coordinates {
    (4.918,1) (4.864,2) (4.882,3)
};
\addlegendentry{Quality}

\addplot+[xbar, fill=Green!90, draw=black] coordinates {
    (4.932,1) (4.858,2) (4.89,3)
};
\addlegendentry{Engagement}

\addplot+[xbar, fill=Red!80, draw=black] coordinates {
    (4.98,1) (4.864,2) (5.000,3)
};
\addlegendentry{Personalness}

\end{axis}
\end{tikzpicture}
\caption{Average evaluation ratings by GPT-4 across gender}
\label{fig:evaluation_gender}
\end{figure}
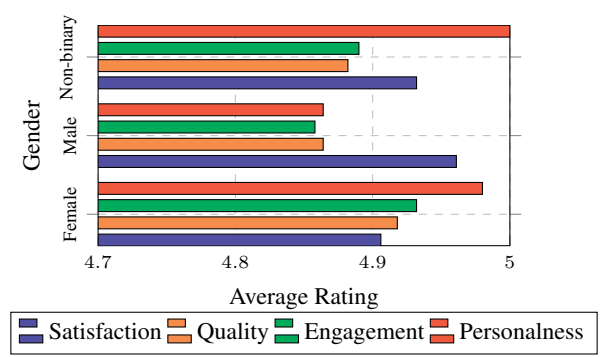

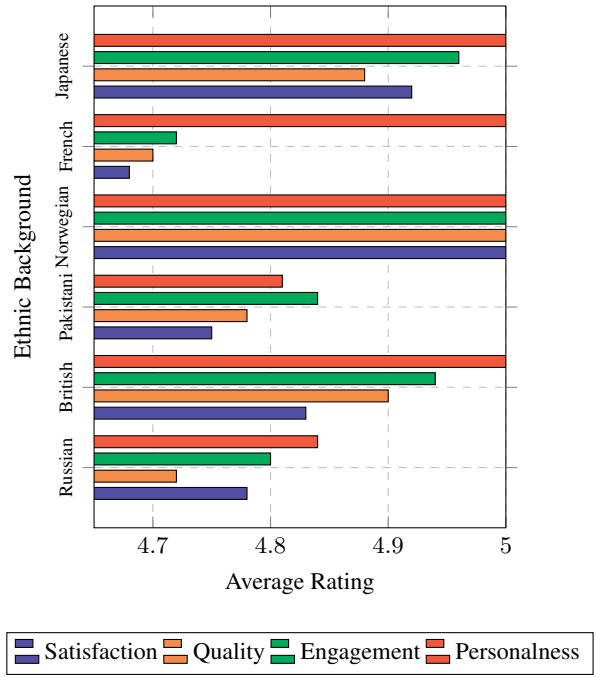
\begin{figure}[h]
\vspace{-2mm}
\centering
\begin{tikzpicture}
\begin{axis}[
    ylabel={Ethnic Background},
    xlabel={Average Rating},
    xlabel style={font=\fontsize{8pt}{10pt}\small},
    ylabel style={font=\fontsize{8pt}{10pt}\small},
    xbar,
    bar width=4.5pt,
    enlarge y limits=0.15,
    ytick=data,
    yticklabels={Russian, British, Pakistani, Norwegian, French, Japanese},
    y tick label style={rotate=90},
    yticklabel style={font=\scriptsize}, 
    xticklabel style={font=\footnotesize}, 
    legend style={at={(0.5,-0.2)},anchor=north, font=\small, legend columns=-1},
    grid=major, % Add major grid lines
    grid style=dashed,
    xmin=4.65, xmax=5.0, 
    width=7cm,
    height=8.5cm
]

\addplot+[xbar, fill=Blue!80, draw=black] coordinates {
    (4.78,1) (4.83,2) (4.75,3) (5, 4) (4.68, 5) (4.92, 6)
};
\addlegendentry{Satisfaction}

\addplot+[xbar, fill=Orange!90, draw=black] coordinates {
    (4.72,1) (4.9,2) (4.78,3) (5, 4) (4.7, 5) (4.88, 6)
};
\addlegendentry{Quality}

\addplot+[xbar, fill=Green!90, draw=black] coordinates {
    (4.8,1) (4.94,2) (4.84,3) (5, 4) (4.72, 5) (4.96, 6)
};
\addlegendentry{Engagement}

\addplot+[xbar, fill=Red!80, draw=black] coordinates {
    (4.84,1) (5,2) (4.81,3) (5, 4) (5, 5) (5.0, 6)
};
\addlegendentry{Personalness}

\end{axis}
\end{tikzpicture}
\caption{Average evaluation ratings by GPT-4 across ethnic background}
\label{fig:evaluation_ethnicity}
\end{figure}

\newpage
% \clearpage
\subsection{Extended Analysis}

\subsubsection{Qualitative Analysis}
\label{appendix_A3_3}

\begin{figure*}[h]
    \centering
    \includegraphics[width=15cm]{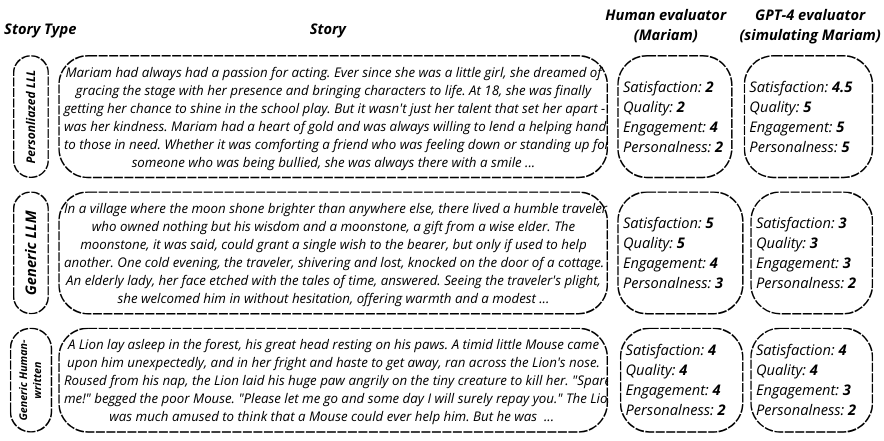}
    \caption{Qualitative comparison of ratings for three types of stories by both human evaluators and GPT-4, including both conflicting and consistent ratings}
    \label{fig:qualitative_comparison}
\end{figure*}

\newpage
\subsubsection{Correlation Analysis}
\label{appendix_A31}

\begin{table}[h]
\centering
\setlength{\tabcolsep}{4.5pt} % Adjusted horizontal padding between columns
\renewcommand{\arraystretch}{1.2}
{\normalsize 
\begin{tabular}{c|cccc} 
\toprule
\textbf{Story Type} & \textbf{Satisfaction} & \textbf{Quality} & \textbf{Engagement} & \textbf{Personalness} \\ 
\midrule
 Personalized LLM-generated & 0.24 & 0.47 & 0.34 & 0.08 \\
 Generic LLM-generated      & 0.19 & 0.22 & 0.20 & 0.12 \\
 Generic Human-written      & 0.12 & 0.26 & 0.21 & 0.19 \\

\bottomrule
\end{tabular}
}
\caption{Spearman's rank correlation coefficients between human evaluators and GPT-4 for story evaluation metrics, where values closer to 1 indicate a stronger positive correlation}
\label{tab:spearman_correlation_results}
\end{table}

% \vspace{2cm}

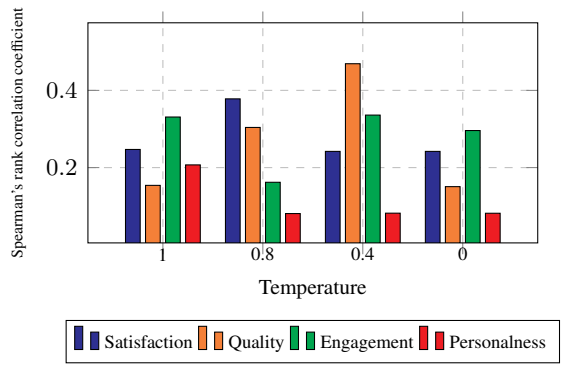
\begin{figure}[h]
\vspace{-1mm}
\centering
\begin{tikzpicture}
\begin{axis}[
    % title=\textbf{\small Temperature vs Scores},
    xlabel={Temperature},
    ylabel={Spearman's rank correlation coefficient},
    ylabel style={font=\fontsize{6pt}{8pt}\selectfont},
    xlabel style={font=\fontsize{8pt}{10pt}\selectfont},
    ybar,
    bar width=5.5pt,
    enlargelimits=0.25,
    xtick=data,
    xticklabels={1, 0.8, 0.4, 0},
    x tick label style={rotate=0,anchor=center},
    xticklabel style={font=\scriptsize}, 
    yticklabel style={font=\footnotesize}, 
    legend style={at={(0.5,-0.35)},anchor=north, font=\scriptsize, legend columns=-1},
    grid=major, % Add major grid lines
    grid style=dashed,
    ymin=0.1, ymax=0.48, 
    width=7.5cm,
    height=4.5cm
]

\addplot+[ybar, fill=Blue, draw=black] coordinates {
    (1,0.247) (2,0.378) (3,0.242) (4,0.242)
};
\addlegendentry{Satisfaction}

\addplot+[ybar, fill=Orange, draw=black] coordinates {
    (1,0.154) (2,0.304) (3,0.469) (4,0.1507)
};
\addlegendentry{Quality}

\addplot+[ybar, fill=Green, draw=black] coordinates {
    (1,0.331) (2,0.162) (3,0.336) (4,0.296)
};
\addlegendentry{Engagement}

\addplot+[ybar, fill=Red, draw=black] coordinates {
    (1,0.207) (2,0.081) (3,0.082) (4,0.0818)
};
\addlegendentry{Personalness}

\end{axis}
\end{tikzpicture}
\caption{Spearman's rank correlation coefficient between human evaluators and GPT-4 at various temperatures for personalized LLM-generated stories}
\label{fig:spearman_temperature}
\end{figure}

% \newpage
\subsubsection{Image generation for personalized stories}
\label{appendix_A34}

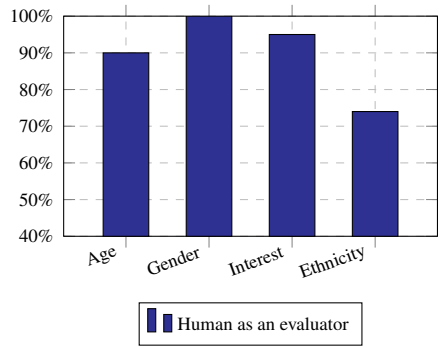
\begin{figure}[h]
\centering
\begin{tikzpicture}
\begin{axis}[
    % title=\textbf{\small Human as an evaluator},
    ylabel style={font=\fontsize{8pt}{10pt}\selectfont},
    xlabel style={font=\fontsize{8pt}{10pt}\selectfont},
    ybar,
    enlarge x limits=0.25,
    bar width=0.6cm,
    xtick=data,
    xticklabels={Age, Gender, Interest, Ethnicity},
    x tick label style={rotate=20,anchor=east},
    xticklabel style={font=\scriptsize}, 
    yticklabel style={font=\scriptsize}, 
    legend style={at={(0.5,-0.3)},anchor=north,font=\scriptsize}, 
    grid=major, 
    grid style={dashed},
    ymin=40, ymax=100, 
    width=6.5cm,
    height=4.5cm,
    ytick={40,50,60,70,80,90,100},
    yticklabels={40\%,50\%,60\%,70\%,80\%,90\%,100\%}
]

\addplot+[ybar, fill=Blue, draw=black] coordinates {
    (1,90) (2,100) (3,95) (4,74)
};
\addlegendentry{Human as an evaluator}

\end{axis}
\end{tikzpicture}
% \\
\caption{Accuracy of human and LLM evaluators in identifying personalization elements in the image}
\label{fig:personalization_test_images_results}
\end{figure}

\begin{figure}[h]
    \centering
    \includegraphics[width=14.5cm]{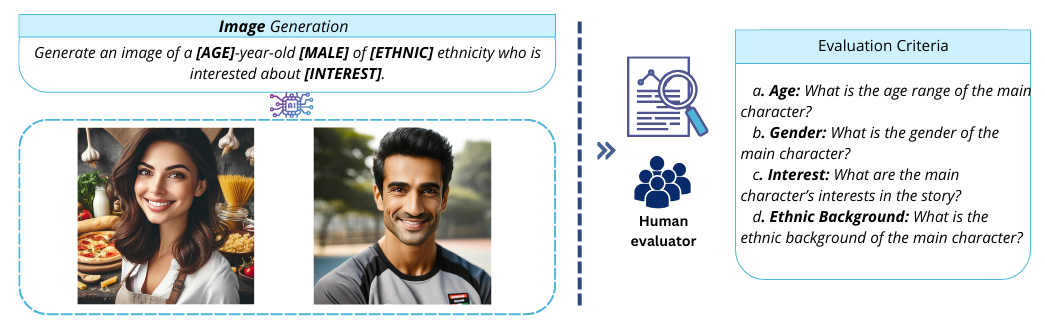}
    \caption{\small Illustration of the personalization test process for images. The left side displays the prompt used to generate personalized images. The right side outlines the evaluation criteria for human evaluators to determine how effectively personal elements have been integrated into the image.}
    \label{fig:personalization_test_images}
\end{figure}

\newpage
\subsection{Dataset Structure, Categories and Values}
\label{sec:subsectionA4}

\begin{table*}[h]
    \centering
    \small
    \begin{tabularx}{\textwidth}{|p{0.65cm}p{0.4cm}p{0.6cm}p{1cm}p{1.0cm}p{1.2cm}|p{5.3cm}|p{1.1cm}|X|} 
        \toprule
        Name &Age&Gender&Interest&Ethnicity& Moral & LLM-generated Personalized Story & LLM-generated Generic Story & Human-written Generic Story \\ 
        \midrule
        Behzad & 23 & Male & Boxing & Indian & Learn from your mistakes & 
        
Behzad, a sturdy 23-year-old Indian man, was far more attached to his boxing gloves than his engineering textbooks. Son of a mechanic, his life was perched comfortably on the rusty edges of Mumbai, lit up by the lambent moon by night and the intruding sun by day. Unlike his peers, Behzad wasn't fascinated by the jingling call of corporate ladders; instead, he ... & \textbf{...} & \textbf{...}
\\
\midrule
Karina &29&Female  &History& Ukrainian& Strive for excellence & Karina's eyes danced with curiosity as she traced her fingers over the ancient tome. The worn leather binding and yellowed pages spoke of centuries past, each word a whisper from a forgotten era. For her, the study of history was more than a mere pursuit of knowledge‚ it was a gateway to understanding ... & \textbf{...} & \textbf{...}
\\ 
\midrule
Dario &35&Non-binary&Volunteering & \hspace{4pt} Italian & Take responsibility for your actions & Dario's eyes sparkled with determination as they stepped into the volunteer center. At 35 years old, they had already made a name for themselves in the local community for their unwavering commitment to making a difference.

Born to Italian parents who instilled a deep sense of responsibility and compassion ... & \textbf{...} & \textbf{...}
\\
        \bottomrule
    \end{tabularx}
    \caption{Dataset structure of \corpusname{}. The dataset includes personal attributes (Name, Age, Gender, Interest, Ethnicity), a moral, and three types of stories: LLM-Generated Personalized Story, LLM-Generated Generic Story, and Human-Written Generic Story.}
    \label{tab:core_dataset_structure}
\end{table*}

\begin{table*}[h]
    \centering
    \normalsize
    
    \begin{tabularx}{\textwidth}{|p{3cm}|X|}
        \hline
        \textbf{Category} & \textbf{Values} \\ 
        \midrule
        Age & 10 ... 60.
        \\
        \midrule
        Gender & Male, Female, Non-binary.
        \\
        \midrule
        Ethnicity&Albanian, Arab, Arab-Amazigh, Armenian, Australian, Austrian, Akan, Andorran, Azerbaijani, Bambara, Belarusian, Bengali, Baganda, Bosnian, Pardo Brazilian, Bosniak, British, Bulgarian, Canadian, Chechen, Chinese, Congolese, Croat, Czech, Dane, Dutch, Egyptian, Emirati, Estonian, Fijian, Finn, Georgian, German, Greek, Hawaiian, Hungarian, Indian, Indonesian (Javanese), Iraqi Arab, Irish, Italian, Japanese, Jewish, Jordanian, Kazakh, Korean, Kikuyu, Kurdish, Kuwaiti, Kyrgyz, Latvian, Lithuanian, Luxembourger, Malay, Maldivian, Maltese, Maori, Mestizo, Moldovan, Norwegian, Punjabi, Palestinian, Persian, Polish, Portuguese, Romanian, Russian, Hutu, Salvadoran, Scottish, Serb, Slovak, Slovene, Somali, Spanish, Sudanese, Swede, Swiss, Syrian, Tajik, Thai, Turk, Turkmen Ukrainian, Uzbek, Vietnamese (Viet), Welsh, Wolof.
        \\
        \midrule
        Interest &Acting, Archery, Arts, Astronomy, Badminton, Bagpiping, Baking, Ballet, Baseball, Basketball, Beadwork, Beekeeping, Biology, Biking, Blogging, Board, Bonsai, Boxing, Calligraphy, Camping, Canoeing, Carpentry, Chess, Coding, Community Service, Cooking, Crafting, Cricket, Culinary, Cycling, Dancing, Digital, Drawing, Drumming, Embroidery, Falconry, Farming, Fashion, Fencing, Filmmaking, Fishing, Football, Foraging, Gardening, Geography, Graphic design, Guitar, Gymnastics, Hiking, History, Hockey, Horseback, Ice, Ikebana, Judo, K-Pop, Kayaking, Kendo, Kickboxing, Kite Flying, Knitting, Literature, Jewelry Making, Martial Arts, Massage, Meditation, Mountaineering, Music, Painting, Papercraft, Parkour, Photography, Piano, Pilates, Poetry, Politics,; Pottery, Quilting, Reading, Respect, Riding, Robotics, Rock Climbing, Rowing, Rugby, Running, Sailing, Sculpting, Science, Sewing, Skateboarding, Skydiving, Skiing, Singing, Social Activities, Soccer, Sprinting, Storytelling, Surfing, Swimming, Taekwondo, Table, Tango, Teaching, ennis, Traveling, Trekking, Video Games, Violin, Volleyball, Volunteering, Weaving, Weightlifting, Winemaking, Woodworking, Wrestling, Writing, Yoga.
        \\
        \midrule
        Moral &Maintain humility, Learn from your mistakes, Be optimistic, Show empathy, Be loyal, Work hard and stay humble, Live with purpose, Take responsibility for your actions, Always tell the truth, Cherish your family, Be generous, Keep your promises, Treat others as you want to be treated, Be a good listener, Be fair and just, Live with integrity, Protect the weak and vulnerable, Seek justice, Be curious and keep learning, Be grateful, Have courage, Help those in need, Strive for excellence, Have respect for yourself and others, Practice good manners, Embrace diversity, The race is not always to the swift, A kindness is never wasted, Liars are not believed even when they speak the truth.
\\
    \bottomrule
    \end{tabularx}
    \caption{Breakdown of the different categories and values included in the \corpusname{} dataset. It covers a diverse range of ages, genders, ethnicities, interests, and moral values.}
    \label{tab:dataset_values}
\end{table*}

\end{document}